\documentclass{article}

\usepackage{times}
\usepackage{microtype}
\usepackage{graphicx}
\usepackage{subfigure}
\usepackage{booktabs} 
\usepackage{multicol}
\usepackage[bookmarks=true]{hyperref}
\usepackage{verbatim} 
\usepackage{algorithm}
\usepackage{algorithmic}
\usepackage{array}
\usepackage{multirow}
\usepackage{amsmath}
\usepackage{amsfonts}
\usepackage{amssymb}
\usepackage{mathrsfs}
\usepackage{bbold}
\usepackage{tikz}
\usepackage{pgfplots}
\usepackage{textcomp}
\pgfplotsset{width=8.0cm,height=4.4cm,compat=1.9}
\usepackage{hyperref}
\newcommand{\mb}[1]{\mathbf{#1}}
\usepackage[accepted]{icml2018}

\icmltitlerunning{MPC-Inspired Neural Network Policies for Sequential Decision Making}
\begin{document}

\twocolumn[
\icmltitle{MPC-Inspired Neural Network Policies for Sequential Decision Making}



\icmlsetsymbol{equal}{*}

\begin{icmlauthorlist}
\icmlauthor{Marcus Pereira}{equal,to}
\icmlauthor{David D. Fan}{equal,to}
\icmlauthor{Gabriel Nakajima An}{to}
\icmlauthor{Evangelos Theodorou}{to}
\end{icmlauthorlist}

\icmlaffiliation{to}{Autonomous Control and Decisions Lab, Georgia Institute of Technology, Atlanta, GA}

\icmlcorrespondingauthor{Marcus Pereira}{mpereira30@gatech.edu}

\icmlkeywords{Machine Learning, ICML, Imitation Learning, Model Predictive Control, Path Integral, DAgger}

\vskip 0.3in
]



\printAffiliationsAndNotice{\icmlEqualContribution} 

\begin{abstract}
In this paper we investigate the use of MPC-inspired neural network policies for sequential decision making.   We introduce an extension to the \textsc{DAgger} algorithm for training such policies and  show how they have improved training performance and generalization capabilities.  We take advantage of this extension to show scalable and efficient training of complex planning policy architectures in continuous state and action spaces.  We provide an extensive comparison of neural network policies by considering feed forward policies, recurrent policies, and recurrent policies with planning structure inspired by the Path Integral control framework.  Our results suggest that MPC-type recurrent policies have better robustness to disturbances and modeling error.
\end{abstract}

\section{Introduction}

Prior work in the area of reinforcement learning utilizes neural network representations to learn  \textit{reactive} policies, i.e. a simple feed-forward neural network, that directly map observations to actions.  Recent work argues the value of introducing planning structure to these policies, in order to improve the ability of these policies to plan, generalize, and learn new tasks \cite{Tamar2016ValueNetworks,Karkus2017QMDP-Net:Observability,Weber2017Imagination-AugmentedLearningb}.  These approaches combine model-based and model-free reinforcement learning into a unified framework where the structure of the policy performs planning, with the individual components of the planner (e.g. dynamics, cost, or value function) learned end-to-end.  Because these planning policies often involve complex architectures, it is easier to evaluate them using imitation learning methods rather than full reinforcement learning algorithms, that require exploration and typically have poor sample efficiency.

The complexity of aformentioned  architectures has thus far restricted them to discrete state and action spaces, where they have been shown to perform well for grid-world based tasks such as planning a path through a maze, Pacman, etc.  To our knowledge, planning policies for continuous state and action spaces have not been adequately addressed yet.  One problem in finding good architectures for  these policies is that for continuous state and action space problems, value function approximation is a difficult problem in its own right.  We hypothesize that trajectory optimization methods, i.e. where a planned trajectory of finite length is iteratively optimized, will be more sample efficient and computationally tractable.  Specifically we propose using Model Predictive Control (MPC)-type policies, in which a sequence of controls is iteratively passed forward and reoptimized at the next time-step.

Model Predictive Control is a well-established method for real-time control which has many advantages, including being robust to disturbances, modeling errors, and suboptimal optimizations \cite{camacho2013model}.  It performs an optimization of future control sequences at each timestep, for which only the first control in the sequence is applied to the system.  At the next timestep the control sequence is carried forward and reoptimized.  This gives the optimizer a "warm-start" which reduces the computational burden placed upon it, allowing for a greater degree of real-time control. In this paper  we leverage the advantages of MPC in our planning policies for three reasons: 1) To reduce the computational burden of optimization required at each timestep; 2) to create a policy which is more robust to disturbances; 3) to improve generalization when the policy has model errors or makes mistakes.
\vspace{-0.1cm}
The main contributions of our work are as follows:
\begin{enumerate}
\item We propose the use of MPC-type policies for sequential decision making and introduce an extension to the \textsc{DAgger} algorithm for training such policies.
\item We take advantage of this extension to show scalable and efficient training of complex planning policy architectures, i.e. PI-Nets.  Our methodology improves time and memory requirements by a factor of 50 when compared with prior work.
\item We provide an extensive comparison of neural network policies on three continuous state and action tasks.  In particular we consider feed forward policies, recurrent policies, and recurrent policies with planning structure, namely, PI-Nets.
\end{enumerate}

\section{Related Work}

Learning dynamics for model-based control and reinforcement learning is a difficult problem, and has been directly addressed in various ways \cite{Venkatraman2017Predictive-StateNetworks,Mishra2017PredictionModels,Ross2012AgnosticLearning}.  Learning planning policies with abstract dynamics end-to-end may circumvent this difficulty.  \cite{Tamar2016ValueNetworks} proposed Value Iteration Networks as a means to embed a value-iteration algorithm in a policy using convolutional layers.  The architecture was restricted to grid world problems with a few discrete actions per state, but showed promising results in terms of generalization.  An extension of their work to the partially observable case was proposed by \cite{Karkus2017QMDP-Net:Observability}, who added Bayesian filtering to the value iteration algorithm.  Parallel to these efforts is the work of \cite{Weber2017Imagination-AugmentedLearningb} on imagination-augmented agents.  They proposed a complex architecture with multiple rollouts of abstract states, which feed into a layer which aggregates information across the multiple rollouts as well as a simple feed forward policy.  Another notable contribution is the work of \cite{farquhar2017treeqn} who propose a tree-structured model which performs rollouts of abstract transition dynamics to estimate the cost-to-go.  \cite{Groshev2017LearningNetworks} propose a planning-type policy which performs A star searches.  All of these methods are applicable to problems with discrete action spaces only, and do not scale well to continuous state and action spaces.  In contrast, \cite{Okada2017PathControl} proposed Path Integral Networks as a way to create a fully differentiable planning policy for continuous state and action spaces.  Unfortunately, the proposed method was computationally prohibitive and does not scale well.  They also failed to demonstrate end-to-end learning of an abstract planning module.  In this work we address these issues with an extension of imitation learning to sequences of controls derived from an MPC expert, and show how this method can be used to efficiently train PI-Net in a fully end-to-end fashion.  We further discuss the differences of our approach with previous work on PI-Nets in section \ref{sec:pinets}.
In this work we assume that the expert is an MPC controller, which gives sequences of controls for a given state.  There has been a wide of range of work on imitation learning which use the same tactic of assuming additional information is available from the expert.  For example, \cite{Ross2014} and \cite{Sun2017DeeplyPrediction} assume an expert is available which gives a cost-to-go.  \cite{Choudhury2017Data-drivenLearning} assume the expert has additional knowledge of the environment, such as a world map.  Most similar to our work is that of \cite{Sun2017a} which use \textsc{DAgger} to train policies to predict a few timesteps of controls from an MPC expert, however, they do not consider training policies to act in an MPC fashion.

Sequence prediction is another area related to our approach, for instance machine translation or generating captions for images \cite{Ranzato2015SequenceNetworks}.  Such networks are trained to produce the next token in the sequence, given the previous token generated by the network.  Naively training a recurrent neural network to do this results in problems such as unstable training and poor performance, for which several approaches have been proposed to address.  For instance, \cite{Bengio2015a} introduced scheduled sampling, which stochastically mixes the data during training, resulting in a \textsc{DAgger}-like algorithm.  \cite{Huszar2015HowAdversary} and \cite{Lamb2016ProfessorNetworks} discuss potential drawbacks of this approach, proposing using various techniques to help the sequence generator match the same distribution of states between training and test time.  While these approaches have a similar flavor to our proposed method, they deal with the problem of sequence generation specifically.  In contrast, our work is concerned with repeatedly optimizing a sequence of controls given the current state of the system.  Borrowing some of the more advanced ideas for training sequence-generating networks and adopting them for our problem merits further investigation; however, in this paper we start with the basic \textsc{DAgger} algorithm and propose an extension to it to directly address the problem at hand.  

\section{Sequential Imitation Learning}
\subsection{Vanilla \textsc{DAgger}}
\begin{figure}[h!]
\centering
\includegraphics[width=0.4\textwidth, angle=0]{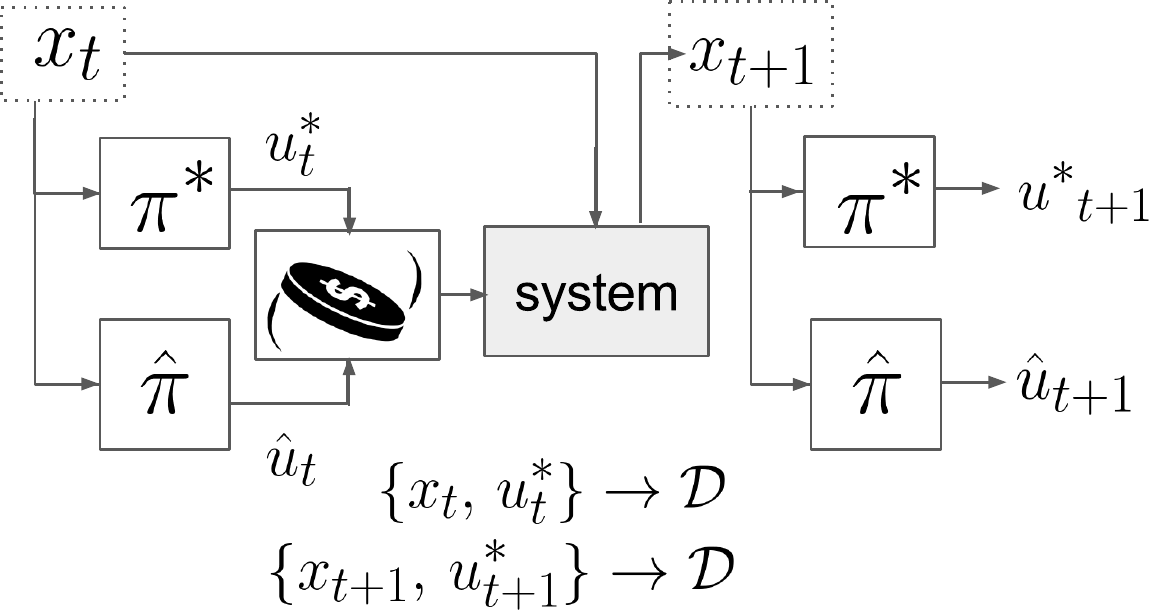}
\caption{Two timesteps of data aggregation using vanilla DAgger}
\label{fig:dagger}
\end{figure}

We first briefly review the vanilla \textsc{DAgger} algorithm then discuss its extension to MPC experts.  Let $x_t\in\mathcal{R}^n$ be the state at time $t=1,\ldots,T$ where $T$ is the task horizon.  Let $\pi\in\Pi$ be a policy within a family of parameterized policies $\Pi$, which produce an action $\pi(x)=u\in\mathcal{R}^m$.  We have a system in which the next state depends on the current state and the policy as $p(x_{t+1}|x_t,\pi(x_t))$.  Let $d_t^\pi$ be the distribution of states when applying $\pi$ to the system for timesteps $1,\ldots, t-1$.  Let $d^\pi=\frac{1}{T}\sum_{t=1}^Td_t^{\pi}$ be the average distribution of states visited over the entire task when following $\pi$. If each state and action pair has an associated cost $C(x,u)$, the total cost-to-go of a policy will be $J(\pi)=\mathbb{E}_{x\sim d^\pi}[C(x,\pi(x))]$.  In imitation learning we may not be provided with this cost function $C$ but instead are provided with an expert which knows how to minimize this cost function.  The goal is to find a policy $\hat{\pi}$ which minimizes a loss function $l(x,\pi)$ that penalizes deviation from an expert policy $\pi^*$, under the distribution of states visited by the policy $\pi$:
\begin{equation}
\hat{\pi}=\arg\min_{\pi\in\Pi}\mathbb{E}_{x\sim d^\pi}[l(x,\pi)].
\end{equation}
Simply training a learner policy on data collected from an expert yields a supervised policy $\hat{\pi}_{sup}$:
\begin{equation}
\hat{\pi}_{sup}=\arg\min_{\pi\in\Pi}\mathbb{E}_{x\sim d^{\pi^*}}[l(x,\pi)].
\end{equation}
Unfortunately, as shown in \cite{Ross2010ALearning}, the cost to go of following this policy trained with supervised learning grows quadratically with task horizon, i.e. if $\mathbb{E}_{x\sim d^{\pi^*}}[l(x,\pi)]=\epsilon$, then $J(\hat{\pi}_{sup})\leq J(\pi^*)+T^2\epsilon$.  \textsc{DAgger} addresses this problem by iteratively training a policy, collecting data with that policy, aggregating the data collected, and retraining a policy for the next iteration.  (See Algorithm \ref{alg:1}, Figure \ref{fig:dagger}).  At each iteration, when applying the newest learned policy to the system, instead of applying the learner policy alone, the expert policy is "mixed" in, i.e. at iteration $i$ the expert policy is applied with probability $\beta_i\in[0,1]$ and the learner policy is applied with probability $(1-\beta_i)$.  The only requirement for the sequence of $\beta_{1:N}$ is that their average decreases to 0, i.e. $\frac{1}{N}\sum_{i=1}^N\beta_i\rightarrow 0$.  The \textsc{DAgger} algorithm has linear regret bounds.  Let $N$ be the number of iterations of \textsc{DAgger}, and let $\epsilon_N=\arg\min_{\pi\in\Pi}\frac{1}{N}\mathbb{E}_{x\sim d^{\pi_i}}[l(x,\pi)]$.  Then Ross et. al. 2011? show that the cost to go is linearly bounded with respect to the time horizon, i.e. if $N$ is $\tilde{O}(aT)$, then there is a policy $\hat{\pi}\in\pi_{1:N}$ such that $J(\hat{\pi})\leq J(\pi^*)+aT\epsilon_N+O(1)$.\\
\subsection{\textsc{DAgger} for MPC Policies}
\begin{figure}[h!]
\includegraphics[width=0.48\textwidth, angle=0]{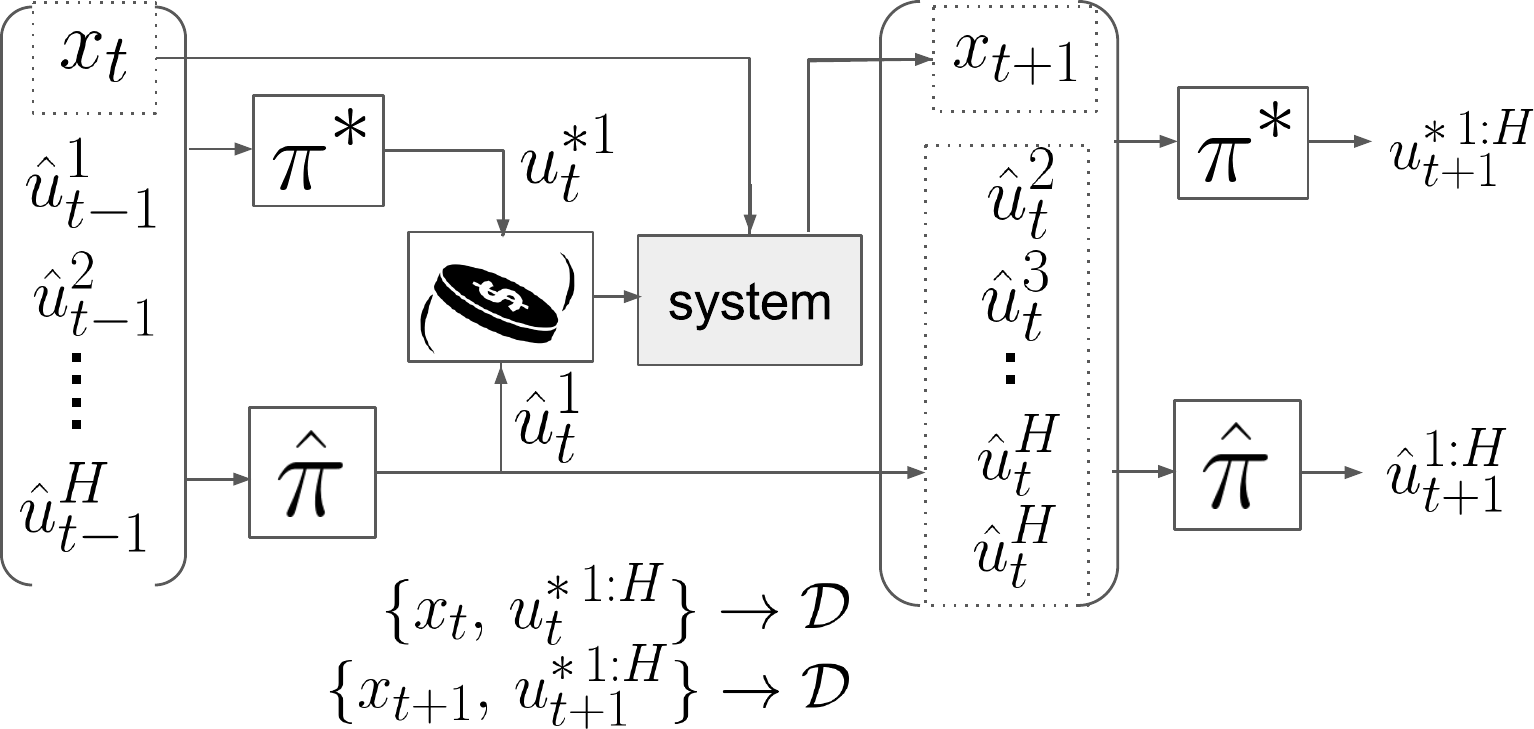}
\caption{Two timesteps of data aggregation using MPC-DAgger}
\label{fig:mpcdagger}
\end{figure}

Next we consider the case when the expert provides a sequence of controls instead of just one control per state.  In this case, by augmenting the state with the control sequence space, we arrive at a similar algorithm and the same theoretical analysis.  We want to learn policies of the form $\pi^{1:H}(x,u^{1:H})$ which produces a sequence of $H$ controls, where $H<<T$ is the prediction horizon and $u^{1:H}$ is some initial sequence of controls provided to the policy as an input.  Let $\Pi'$ be the family of policies of this form.  We augment the state space with $H$ control inputs, as $\bar{x}=[x,u^{1:H}]\in\mathcal{R}^n\times{(\mathcal{R}^m)}^H$, so we can write $\pi^{1:H}(\bar{x})\in\Pi'$.  When moving from one timestep to the next, we assume that that policy uses its own output from the previous timestep as a "warm start", by shifting the control sequence by one and padding the end with the last control $u^{1:H}=[\pi^{2:H}(\bar{x}),\pi^H(\bar{x})]$.  In this way we have fully deterministic dynamics in the augmented control space, and the same dynamics in the state space as before, i.e. $p(x_{t+1}|x_t,\pi^1(x_t,u_t^{1:H}))$.  (Figure \ref{fig:mpcdagger}).  We can then define $\bar{d}_t^\pi$ as the distribution of augmented states when following $\pi\in\Pi'$ for $t-1$ timesteps, and $\bar{d}^\pi=\frac{1}{T}\sum_{t=1}^Td_t^\pi$ as the average distribution of augmented states visited over the entire task.  Let $\bar{l}(\bar{x},\pi)$ be a loss function of the augmented state and the policy which penalizes deviation of the entire control trajectory with that of the expert's.  We then have the same problem formulation as vanilla \textsc{DAgger} with an augmented state:
\begin{equation}
\hat{\pi}=\arg\min_{\pi\in\Pi'}\mathbb{E}_{\bar{x}\sim \bar{d}^\pi}[\bar{l}(\bar{x},\pi)].
\end{equation}
We can then construct a similar \textsc{DAgger} algorithm for MPC-type policies given an expert that provides sequences of controls (Algorithm \ref{alg:2}).  Similar theoretical analysis of the \textsc{DAgger} algorithm follows, yielding similar bounds on the cost-to-go, i.e. there is a policy $\hat{\pi}\in\pi_{1:N}$ such that $J(\hat{\pi})\leq J(\pi^*)+aT\bar{\epsilon}_N+O(1)$, where $\bar{\epsilon}_N=\arg\min_{\pi\in\Pi'}\frac{1}{N}\mathbb{E}_{\bar{x}\sim \bar{d}^{\pi_i}}[\bar{l}(\bar{x},\pi)]$.\\
Since \textsc{DAgger} calls for querying the expert on the distribution of states that the learner sees, and in this case  the state is augmented with the control sequences of the learner, this means that we must provide the expert with the learner's control sequences to reoptimize, i.e. $\pi^{*1:H}(x,[\hat{\pi}^{2:H}, \hat{\pi}^H])$.
However, in practice we found that warm-starting the expert with its own control sequence yielded better results than having the expert reoptimize the learner's output control sequence.  This change can be justified by assuming the expert depends only on the current state and not the warm-starting, i.e $\pi^{*1:H}(x,[\hat{\pi}^{2:H}, \hat{\pi}^H])=\pi^{*1:H}(x)$.

\begin{algorithm}
\label{alg:1}
\caption{Vanilla \textsc{DAgger}}
\begin{algorithmic}
\REQUIRE \textsc{DAgger} iterations N, Episode length T, Expert $\pi^*$, Horizon $H$, Beta schedule $\beta_{1:N}$
\STATE Initialize dataset $\mathcal{D}\leftarrow\{\emptyset\}$.
\STATE Initialize learner policy $\hat{\pi}_1$ arbitrarily.
\FOR{$i=1$ \TO $N$}
\STATE Initialize initial state $x_1$.
\FOR{$t=1$ \TO $T$}
\STATE Query expert and learner controls:
\STATE $\pi^*(x_t)$, $\hat{\pi}_i(x_t)$
\IF {$\mathcal{U}[0,1] < \beta_i$}
\STATE Apply expert's control $\pi^*(x_t)$ to system.
\ELSE
\STATE Apply learner's control $\hat{\pi}(x_t)$ to system.
\ENDIF
\STATE Add data point: $\mathcal{D}\leftarrow\mathcal{D}\cup\{x_t||\pi^*\}$.
\ENDFOR
\STATE Train policy $\hat{\pi}_{i+1}$ on $\mathcal{D}$.
\ENDFOR
\STATE Return best $\hat{\pi}_i$ in $i=1,\ldots, N$ for testing.
\end{algorithmic}
\end{algorithm}

\begin{algorithm}
\label{alg:2}
\caption{\textsc{DAgger} for MPC Policies}
\begin{algorithmic}
\REQUIRE \textsc{DAgger} iterations N, Episode length T, Sequential expert $\pi^*$, Horizon $H$, Beta schedule $\beta_{1:N}$
\STATE Initialize dataset $\mathcal{D}\leftarrow\{\emptyset\}$.
\STATE Initialize learner policy $\hat{\pi}_1$ arbitrarily.
\FOR{$i=1$ \TO $N$}
\STATE Initialize initial state $x_1$.
\STATE Initialize expert's input control sequence $u^{*1:H}$
\STATE Initialize learner's input control sequence $\hat{u}^{1:H}$
\FOR{$t=1$ \TO $T$}
\STATE Query expert and learner control sequences:
\STATE $\pi^{*1:H}(x_t,u^{*1:H})$, $\hat{\pi}_i^{1:H}(x_t,\hat{u}^{1:H})$
\IF {$\mathcal{U}[0,1] < \beta_i$}
\STATE Apply first element of expert's control sequence $\pi^*_1(x_t,u^{*1:H})$ to system.
\ELSE
\STATE Apply first element of learner's control sequence $\hat{\pi}_1(x_t,\hat{u}^{1:H})$ to system.
\ENDIF
\STATE Add data point: $\mathcal{D}\leftarrow\mathcal{D}\cup\{x_t,\hat{\pi}_i^{1:H}||\pi^{*1:H}\}$.
\STATE Shift and update input control sequences:
\STATE $u^{*1:H}\leftarrow[\pi^{*2:H},\pi^{*H}]$
\STATE $u^{1:H}\leftarrow[\hat{\pi}^{2:H},\hat{\pi}^{H}]$
\ENDFOR
\STATE Train policy $\hat{\pi}_{i+1}$ on $\mathcal{D}$.
\ENDFOR
\STATE Return best $\hat{\pi}_i$ in $i=1,\ldots, N$ for testing.
\end{algorithmic}
\end{algorithm}

\section{PI-Net: End-to-end Differentiable Path Integral Control}
PI-Net \cite{Okada2017PathControl} is a fully differentiable model predictive control algorithm based on the Model Predictive Path Integral (MPPI) controller \cite{Williams2017InformationLearning}.  MPPI is a sampling-based MPC algorithm which finds an analytic expression for the optimal control by sampling over trajectory paths.  It has been applied with success in various domains including aggressive driving tasks.  MPPI does not use any analytic derivatives of the dynamics or cost, in contrast to other trajectory optimization methods such as iLQG \cite{todorov2005generalized}.  This is an important consideration for designing fully differentiable planning architectures, since we would like to avoid taking higher order gradients.  Here we present a brief overview of the MPPI controller and show how it is fully differentiable with respect to the dynamics, cost, and parameters, without invoking these higher order gradients.

Consider the stochastic dynamical system with state and controls $\mathbf{x}_t\in\mathbb{R}^n$ and $\mathbf{u}_t\in\mathbb{R}^m$, and Brownian noise $\text{d}\mathbf{w}\in\mathbb{R}^p$.  The dynamics of the system are described by the control-affine stochastic differential equation:
\begin{equation}
\label{eq:dynamics}
\text{d}\mathbf{x}=\mathbf{f}(\mathbf{x}_t,t)\text{d}t+\mathbf{G}(\mathbf{x}_t,t)\mathbf{u}_t\text{d}t+\mathbf{B}(\mathbf{x}_t,t)\text{d}\mathbf{w}.
\end{equation}
Note that for ease of notation we first formulate the problem in continuous time.  Later we will discretize time to give the full algorithm.
Let $\mathbf{u}(\cdot):[t_0,T]\rightarrow\mathbb{R}^m$ be the control sequence which maps time to control inputs.  We want to find the control sequence $\mathbf{u}(\cdot)$ which minimizes the total cost to go.  The running cost is defined as:
\begin{equation}
\mathbf{L}(\mathbf{x}_t,\mathbf{u}_t,t)=q(\mathbf{x}_t,t)+\frac{1}{2}\mathbf{u}_t^{\intercal}\mathbf{R}(\mathbf{x}_t,t)\mathbf{u}_t.
\end{equation}
The optimal control we want to find is:
\begin{equation}
\label{eq:oc}
\mathbf{u^*}(\cdot)=\underset{\mathbf{u}(\cdot)}{\text{arg}\,\text{min}}\mathbb{E}_\mathbb{Q}\big[\int_{t_0}^H\mathbf{L}(\mathbf{x}_t,\mathbf{u}_t,t)\text{d}t\big],
\end{equation}
where the expectation $\mathbb{E}_\mathbb{Q}$ is taken over trajectories governed by the dynamics described by (\ref{eq:dynamics}).
Using the information theoretic notions of free energy and relative entropy, Model Predictive Path Integral (MPPI) control finds an analytical expression for the optimal control \cite{Theodorou2015}.  Let $(\Omega,\mathscr{F}, \mathbb{P})$ be a probability space where $\Omega$ is the set of all possible trajectories of the state $\mathbf{x}_t$ from time $t\in[t_0,H]$, $\mathscr{F}$ is the $\sigma$-algebra induced by $\Omega$, and $\mathbb{P}$ is the probability measure over trajectories induced by the uncontrolled stochastic dynamics: $\text{d}\mathbf{x}=\mathbf{f}(\mathbf{x}_t,t)\text{d}t+\mathbf{B}(\mathbf{x}_t,t)\text{d}\mathbf{w}$.  Let $\mathbb{Q}$ be a second, arbitrary probability measure which is absolutely continuous with respect to $\mathbb{P}$ (i.e. $\forall A\in \mathscr{F}$ s.t. $\mathbb{P}(A)=0$ we have that $\mathbb{Q}(A)=0$).  Let $\mathbb{Q}^*$ be the optimal probability measure that corresponds to minimizing (\ref{eq:oc}), i.e. the probability measure over trajectories induced by the stochastic dynamics with the optimal control $\mathbf{u}^*(\cdot)$ applied.  One can minimize the KL divergence between the optimal probability measure $\mathbb{Q}^*$ and the probability distribution induced by some non-optimal controller $\mathbb{Q}(\mathbf{u})$:
\begin{equation}
\label{eq:kl}
\mathbf{u}^*(\cdot)=\underset{\mathbf{u}(\cdot)}{\text{arg}\,\text{min}}\mathbb{D}_{KL}(\mathbb{Q}^*\|\mathbb{Q}(\mathbf{u})).
\end{equation}
Since we apply the algorithm in discrete time we can parameterize the control $\mathbf{u}(\cdot)$ as a step function, with $\mathbf{u}(t)=\mathbf{u}_t$, for $t=0,1,\ldots,H$, where we slightly abuse the notation to have $H$ be the number of steps in the control horizon.  We make the following assumptions on $\mathbf{G}$ and $\mathbf{B}$:
\begin{equation}
\label{eq:assumption}
\mathbf{G}=\begin{pmatrix}0\\\mathbb{1}\end{pmatrix},\mathbf{B}(\mathbf{x}_t)=\begin{pmatrix}\mathbf{B}_a(\mathbf{x}_t) & 0\\0 & \nu\mathbb {1}\end{pmatrix},
\end{equation}
where $\nu\geq 1$ controls the amount of exploration.  The minimization can then be solved analytically by taking trajectory-long samples of the expectations in \ref{eq:kl}.  Suppose that we sample $K$ discrete trajectories for $k=1,\dotsc, K$, and let $\tau_k$ be the $k^{\text{th}}$ trajectory, $\tau_k=\{\mathbf{x}_t,\mathbf{u}_t\}_{t=1}^H$.  When sampling these trajectories in a time-discrete manner, the Brownian noise term $\text{d}\mathbf{w}$ in (\ref{eq:dynamics}) becomes $\epsilon_{t,k}\sqrt{\Delta t}$, where $\epsilon_{t,k}\sim\mathcal{N}(0,1)$.  Then the optimal control is approximated by the closed-form expression \cite{RN10}:
\begin{equation}
\label{eq:finalcontrol}
\mathbf{u}_t^{*}=\mathbf{u}_t+\sum_{k=1}^{K}\Bigg(\frac{\exp\big(-\frac{1}{\lambda}\tilde{S}(\tau_k)\big)\frac{\epsilon_{t,k}}{\sqrt{\Delta t}}}{\sum_{k=1}^K\exp\big(-\frac{1}{\lambda}\tilde{S}(\tau_k)\big)}\Bigg),
\end{equation}
where $\mathbf{u}_t$ is the baseline control used to sample the trajectories $\tau_k$ and where the control-cost adjusted running cost $\tilde{S}(\tau_k)$ is
\begin{equation}
\tilde{S}(\tau_k)=\sum_{j=1}^H\tilde{q}(\mathbf{x}_{j},\mathbf{u}_{j},\epsilon_{j,k},j)\Delta t,
\end{equation}
with
\begin{multline}
\label{eq:cost}
\tilde{q}(\mathbf{x}_{j},\mathbf{u}_{j},\epsilon_{j,k},j)=q(\mathbf{x}_{j},j)\\
+\frac{1}{2}\mathbf{u}_j^\intercal\mathbf{R}\mathbf{u}_{j}+\lambda\mathbf{u}_j^\intercal\frac{\epsilon_{j,k}}{\sqrt{\Delta t}}+\frac{\lambda(1-\nu^{-1})}{2}\frac{\epsilon_{j,k}^\intercal\epsilon_{j,k}}{\Delta t}.
\end{multline}
This expression (\ref{eq:finalcontrol}) for the optimal controls can be interpreted as an update rule which computes the new control as a cost-weighted average over the sampled trajectories, given an initial control sequence from which to sample.  Each sampled trajectory is sampled from \ref{eq:dynamics} and \ref{eq:assumption} with controls $\mathbf{u}_t$. The expression is fully differentiable and free of derivatives with respect to the dynamics or cost, making it a favorable candidate for a planning policy architecture.  
\begin{figure*}[!ht]
\includegraphics[width=1.0\textwidth, angle=0]{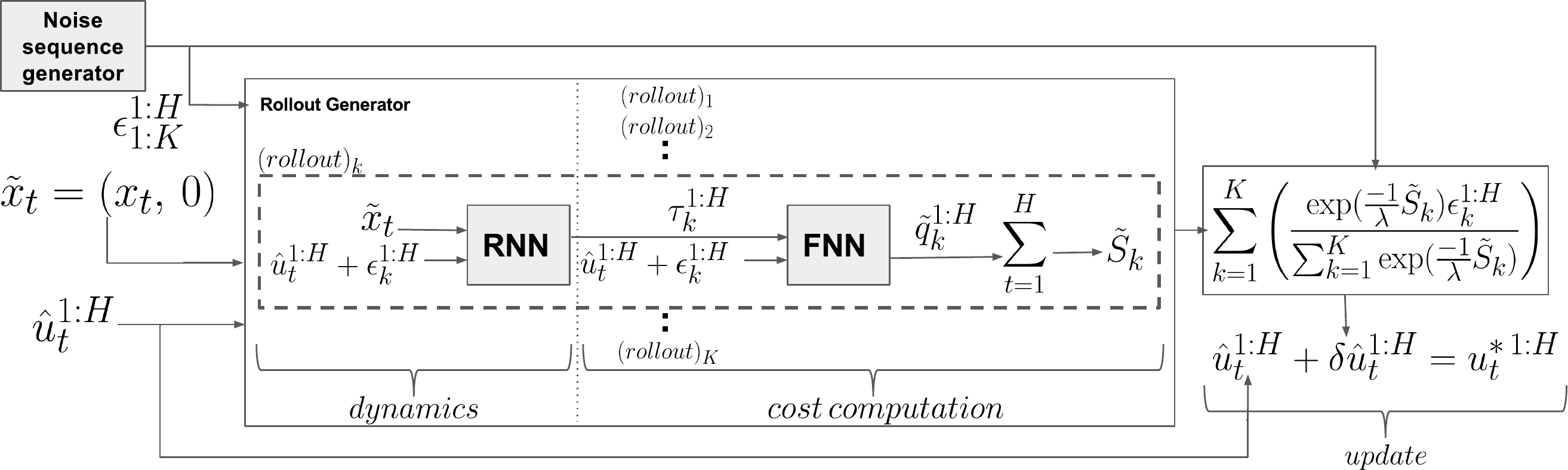}
\caption{PI-Net Architecture}
\label{fig:pinets}
\end{figure*}

\subsection{PI-Net: A Fully Differentiable Optimal Controller}
\label{sec:pinets}
By parameterizing the dynamics $\mathbf{f}(\mathbf{x}_t,t)$ and cost $\tilde{q}(\mathbf{x}_{j},\mathbf{u}_{j},\epsilon_{j,k},j)$ with neural networks, we arrive at a fully differentiable architecture for model predictive optimal control (Figure \ref{fig:pinets}).  The PI-Net architecture can be broken down into the following modules:

\textbf{Dynamics module}: This module consists of a recurrent neural network (RNN) that generates trajectories of an abstract hidden state for $H$ timesteps ahead. The hidden state is initialized by augmenting the input vector $x_t$ (i.e. current state of the system) with zeros. This module generates $K$ samples of abstract state trajectories given the current state of the system, the initial sequence of controls and $K$ sequences of Gaussian random noise. By performing an element-wise addition of the initial control sequence with each of the noise sequences, $K$ perturbed control sequences are generated. The RNN uses these as inputs to generate $K$ abstract state trajectories in parallel. 

\textbf{Cost computation module}: This module comprises of a fully connected feed-forward neural network (FNN), that computes the cost associated with the abstract state trajectories and the corresponding perturbed control sequences, i.e. approximating equation (\ref{eq:cost}) for $\tilde{q}(\mathbf{x}_{j},\mathbf{u}_{j},\epsilon_{j,k},j)$. The output of this layer is then summed over $H$ timesteps.  

\textbf{Update module}: This module performs the \textit{control-update} step given by equation (\ref{eq:finalcontrol}). It receives the input control sequence, the $K$ Gaussian noise signal sequences and the total trajectory costs $\tilde{S_k}$ generated by the cost computation module.

Below we list our contributions and the major differences in our approach to implement the PI-Net architecture as compared to the work by \cite{Okada2017PathControl}: 

\textbf{Scalability}: The original implementation of PI-Nets required evaluating the \textit{dynamics} network $f$ and the cost network $\tilde{q}$, $U\times H \times K \times B$ times during back-propagation and storing all of these values, where $U$ is the number of PI-Net iterations (forward-passes) for the same current state and $B$ is the mini-batch size. Their experiments used over 100GB of RAM which forced training on CPU instead of GPU. In more recent work \cite{Okada2017AccelerationControl}, they introduce a version of PI-Nets with momentum and show that doing so allows them to reduce the value of U (from 200 to 50, 32.5GB of RAM required).  In our approach, $U=1$ which allows us to train PI-Nets for higher dimensional systems on a single GPU.  This reduces both the training time and amount of computational resources required for training as well as makes our approach more feasible for real-time implementation.

\textbf{End-to-end imitation learning for high dimensional non-linear systems}: In \cite{Okada2017PathControl}, the experiments involve training a PI-Nets model end-to-end for a 1-D linear system and separately training the dynamics and cost networks of a PI-Nets model for an inverted pendulum swing-up task. In our experiments, we train our PI-Nets models completely end-to-end (simultaneously learning the dynamics and cost networks) for highly non-linear and under-actuated systems such as cart-pole and quadcopter. Because we train our PI-Nets models end-to-end and use an abstract representation of the original state, the weights of the dynamics and cost networks become correlated. We hypothesize that this helps PI-Nets generalize better than separately training the two networks.

\textbf{RNN dynamics network vs FNN dynamics network}: We demonstrate the ability of our approach to learn an abstract dynamics network (which is itself an RNN) end-to-end along with the cost network to directly sample trajectories, as compared to an FNN trained separately for dynamics in \cite{Okada2017PathControl}.

\section{Experiments}
The PI-Nets architecture comprises of a recurrent neural network (RNN) to capture the system's dynamics and a feed-forward neural network (FNN) to capture the cost (or reward) function latent in the expert's demonstrations. We performed an ablative analysis on the PI-Net architecture by considering the RNN and the FNN as independent policies and recorded the performance of these policies for imitation leaning using MPC-DAgger. We compared the performance of these policies to those trained using \textit{Vanilla} DAgger as well as naive supervised learning. In order to ensure that all policies use approximately the same number of trainable weights and biases, we vary the number of hidden units of each policy to be consistent.  All networks used $tanh()$ activations and with the exception of PI-Net were trained with 500 epochs for each iteration of \textsc{DAgger}. We used the following schedule for $\beta_i=[1.0,\, 0.8,\, 0.7,\, 0.6, \, 0.5, \,  0.45, \, 0.4, \, 0.35, \, 0.30, \, 0.25, \, 0.2, \\0.18, \, 0.16, \, 0.14, \, 0.12, \, 0.10, \, 0.08, \, 0.06, \, 0.04, \, 0.02, \, 0.0,\\ 0.0, \, 0.0]$. 
\subsection{Other Policies}
\textbf{MPC-\textsc{DAgger}: PI-Net}  For our experiments, the RNN and FNN of the PI-Nets model each comprised of a single hidden layer with 64 hidden units.  The horizon $H$ was set to $20$ timesteps, number of PI-Nets iterations $U=1$ and $K=100$ during training. During each iteration of MPC-DAgger, the PI-Nets model was trained on the latest aggregated dataset for 100 epochs. At test-time, it was observed that the performance of the PI-Nets model improved significantly as we increased the number of rollouts to $2000, \, 4000, \, 6000$ etc. and the number of PI-Nets iterations for the same state input to $U=2$. A single GPU was to perform these rollouts at test-time. Thus, our approach allows for training with fewer rollouts during training and increasing the number of rollouts and iterations during test-time, saving computation hours during training.  We trained PI-Net for 100 epochs at each iteration.  Training the full \textsc{MPC-DAgger} algorithm took approximately 3 days.

\textbf{MPC-\textsc{DAgger}: RNN}  We trained an RNN policy that directly outputs the control sequence given the current state of the system and an initial control sequence. As shown in Figure \ref{fig:mpcdagger}, the control sequence predicted by the RNN at the previous timestep is used to initialize the control sequence at the current time step. For our experiments we considered a single hidden layer with 64 hidden units and $tanh()$ activations and a fully connected output layer with units equal to the number of control actions and linear activations. The model was trained on the latest aggregated dataset for 500 epochs during each iteration of MPC-DAgger. 

\textbf{MPC-\textsc{DAgger}: FNN}  This policy is an FNN that takes the same input as the two policies above, except that the input vector is a concatenation of the current state of the system and the initial control sequence. The output of the network is an entire sequence of control actions.

\textbf{Vanilla \textsc{DAgger}: RNN}  Because this policy is trained with Vanilla \textsc{DAgger}, there is no control sequence input, rather the input is a sequence of the states that the system visits when sampling trajectories during each DAgger iteration. The input sequence grows to a maximum length of the number of timesteps in the task.

\textbf{Vanilla \textsc{DAgger}: FNN}  We trained an FNN \textit{reactive} policy that takes the current system state as input and outputs the actions to be applied. 

\subsection{Tasks}
We tested the trained policies on a quadcopter model and a cart-pole model and recorded the performance of each policy (refer Table \ref{tab:1}) with baseline parameters (i.e. the system parameters same as those used during training) and with perturbed parameters to test for robustness.  
\subsubsection{Quadcopter Trajectory Tracking}
We tested the trained policies on two trajectory tracking tasks for a quadcopter model: figure-of-8 and circle.  The expert was MPC-iLQG.  During training, for every \textsc{DAgger} iteration, 64 points were randomly sampled from the set of points on the target trajectory as initial states.  Each episode length was set to $3s$ with a discretization timestep, $dt=0.02s$.  For testing, the length of each episode was increased to $15s$. This episode length was empirically determined as the time required by the expert to complete 1 lap of the figure-of-8 and circle target trajectories. The policies were tested on 4 variations of the trajectory tracking tasks to test for robustness: increased variance additive noise in the controls, randomly perturbing initial states in position, reducing quadcopter arm length, and increasing the quadcopter mass. Costs were calculated by finding the minimum total distance of the generated trajectory from the target trajectory.  A cost over 100 was considered a failure, since these trajectories usually veered wildly away from the target trajectory.  
\subsubsection{Cart Pole Swing-up}
This system comprises of a pole attached to a cart by an un-actuated joint and the cart moves along a rail with some friction. The cart-pole system was randomly initialized as follows: initial cart position is drawn from a uniform distribution $\mathcal{U}[x_{min},x_{max}]$, where $x_{min}=-5.0$ and $x_{max}=5.0$ and the initial pole angular displacement is drawn from a uniform distribution $\mathcal{U}[0,2\pi]$. For this task, MPPI was used as the expert. We performed the \textsc{MPC-DAgger} and \textsc{DAgger} iterations with a batch of $64$ MPPI experts running in parallel for an episode length of $5s$ and discretization timestep, $dt=0.05s$. For testing, we considered 3 generalization tasks: increasing variance of the additive noise, increasing cart mass ($M$), and increasing pole length ($l$). The success-rates and mean scores were calculated based on the distance to the upright centered position of the cart-pole system for the 2nd half of the trials.  Losses over 100 were considered failures, these trajectories failed to swing up and stay upright.

\begin{figure}[!h]
\begin{tikzpicture}
  \begin{axis}[
      xlabel={Mass [$kg$]},
      xlabel style = {font=\footnotesize, yshift=0.5ex,}, 
      ylabel={Success Rate [$\%$]},
      ylabel style = {font=\footnotesize}, 
      xmin=1.00, xmax=1.50,
      ymin=-10, ymax=110,
      xtick={1.0, 1.1, 1.2, 1.3, 1.4, 1.5},
      ytick={0,20,40,60,80,100},
      legend style={legend columns=-1, at={(-0.1,-0.45)},anchor=west},
      ticklabel style = {font=\tiny}, 
      ymajorgrids=true,
      grid style=dashed,
  ]

  \addplot[ color=blue, mark=*]
      coordinates {
      (1, 100.0) (1.05, 100.0) (1.1, 100.0) (1.15, 100.0) (1.2, 100.0) (1.25, 100.0) (1.3, 100.0) (1.35, 97.7) (1.4, 95.3) (1.45, 98.4) (1.5, 96.9) };
      \addlegendentry{\tiny PI-Net}
      
  \addplot[ color=green, mark=triangle]
      coordinates {
      (1.0, 100.0)(1.05, 100.0)(1.1, 100.0)(1.15, 100.0)(1.2, 71.9)(1.25, 47.7)(1.3, 14.8)(1.35, 9.4)(1.4, 0.0)(1.45, 0.0)(1.5, 0.0) };
      \addlegendentry{\tiny MPC-RNN}
      
  \addplot[ color=red, mark=x]
      coordinates {
      (1.0, 100.0)(1.05, 93.0)(1.1, 94.5)(1.15, 83.6)(1.2, 83.6)(1.25, 67.1)(1.3, 56.3)(1.35, 59.4)(1.4, 57.8)(1.45, 46.9)(1.5, 43.8)
      };
      \addlegendentry{\tiny MPC-FNN}
      
  \addplot[ color=brown, mark=square]
      coordinates {
      (1.0, 100.0)(1.05, 100.0)(1.1, 0.0)(1.15, 0.0)(1.2, 0.0)(1.25, 0.0)(1.3, 0.0)(1.35, 0.0)(1.4, 0.0)(1.45, 0.0)(1.5, 0.0) };
      \addlegendentry{\tiny RNN}
      
  \addplot[ color=black, mark=o]
      coordinates {
      (1.0, 100.0)(1.05, 22.7)(1.1, 0.0)(1.15, 0.0)(1.2, 0.0)(1.25, 0.0)(1.3, 0.0)(1.35, 0.0)(1.4, 0.0)(1.45, 0.0)(1.5, 0.0)
      };
      \addlegendentry{\tiny FNN}
  \end{axis}
  
\end{tikzpicture}
\vspace*{-3mm}
\caption{Quadcopter Figure-of-8, Increasing Mass} \label{fig:pinets_figof8}
\end{figure}
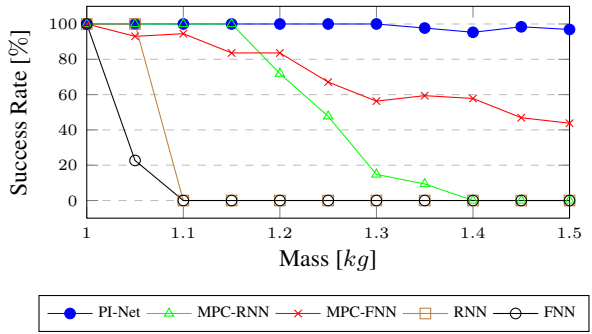

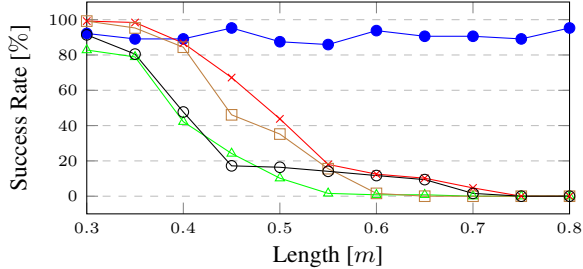
\begin{figure}[!h]
\begin{tikzpicture}
  \begin{axis}[
      xlabel={Length [$m$]},
      xlabel style = {font=\footnotesize, yshift=0.5ex}, 
      ylabel={Success Rate [$\%$]},
      ylabel style = {font=\footnotesize}, 
      xmin=.3, xmax=0.8,
      ymin=-10, ymax=110,
      xtick={0.3, 0.4, 0.5, 0.6, 0.7, 0.8},
      ytick={0,20,40,60,80,100},
      ticklabel style = {font=\tiny}, 
      legend style={at={(1.01,0.5)},anchor=west},
      ymajorgrids=true,
      grid style=dashed,
  ]

  \addplot[ color=blue, mark=*]
      coordinates {
      (0.3, 92.2)(0.35, 89.1)(0.4, 89.1)(0.45, 95.3)(0.5, 87.5)(0.55, 85.9)(0.6, 93.8)(0.65, 90.6)(0.7, 90.6)(0.75, 89.1)(0.8, 95.3)
      };
      
  \addplot[ color=green, mark=triangle]
      coordinates {
      (0.3, 82.8)(0.35, 78.9)(0.4, 42.2)(0.45, 24.2)(0.5, 10.2)(0.55, 1.6)(0.6, 0.8)(0.65, 0.8)(0.7, 0.0)(0.75, 0.0)(0.8, 0.0)
      };
      
  \addplot[ color=red, mark=x]
      coordinates {
      (0.3, 99.2)(0.35, 98.4)(0.4, 86.7)(0.45, 67.2)(0.5, 43.8)(0.55, 18.0)(0.6, 12.5)(0.65, 10.2)(0.7, 4.7)(0.75, 0.0)(0.8, 0.0)
      };
      
  \addplot[ color=brown, mark=square]
      coordinates {
      (0.3, 99.2)(0.35, 95.3)(0.4, 84.4)(0.45, 46.1)(0.5, 35.2)(0.55, 15.6)(0.6, 1.6)(0.65, 0.0)(0.7, 0.0)(0.75, 0.0)(0.8, 0.0)
      };
      
  \addplot[ color=black, mark=o]
      coordinates {
      (0.3, 91.4)(0.35, 80.5)(0.4, 47.7)(0.45, 17.2)(0.5, 16.4)(0.55, 14.1)(0.6, 11.7)(0.65, 9.4)(0.7, 1.6)(0.75, 0.0)(0.8, 0.0)
      };
  \end{axis}
  
\end{tikzpicture}
\vspace*{-4mm}
\caption{Cartpole, Increasing Length} 
\label{fig:cartpole_pinets}
\end{figure}

\begin{figure}[!h]
\centering
\includegraphics[width=0.37\textwidth, angle=0]{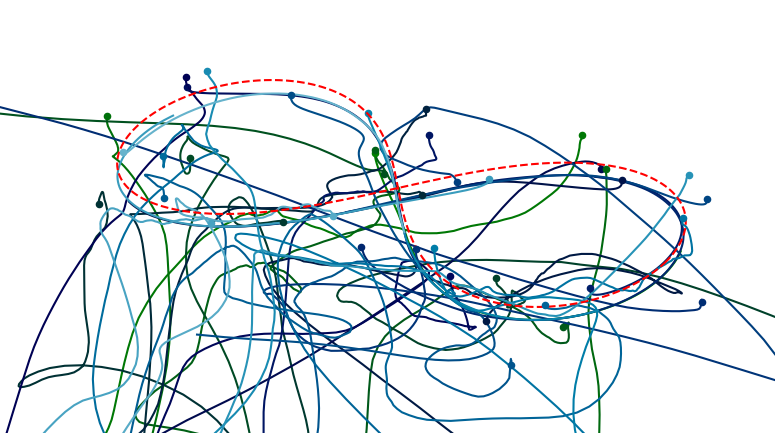}
\includegraphics[width=0.37\textwidth, angle=0]{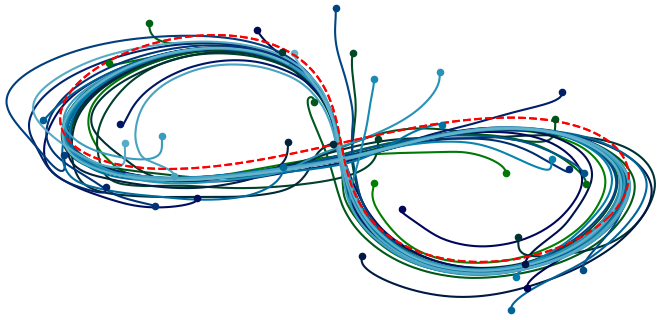}
\caption{Quadcopter Figure-of-8 trajectories with FNN (top) and MPC-RNN (bottom) with perturbed initial positions.  Target trajectory in dotted red.}
\label{fig:fnn_quad}
\end{figure}

\begin{table*}[!t]
  \caption{Comparison of various neural network architectures and learning methods on generalization tasks, showing task success rates and costs (mean and standard deviation).  Costs are computed from successful trials only.}
  \begin{center}
  \begin{tabular}{>{\centering}p{6mm}>{\centering}p{11mm}
  >{\centering}p{4.5mm}>{\centering}p{12mm}
  >{\centering}p{4.5mm}>{\centering}p{12mm}
  >{\centering}p{4.5mm}>{\centering}p{12mm}
  >{\centering}p{4.5mm}>{\centering}p{12mm}
  >{\centering}p{4.5mm}>{\centering}p{12mm}
  >{\centering}p{4.5mm}c}
  \toprule
  \multicolumn{2}{c}{} & \multicolumn{6}{c}{\textbf{\textsc{MPC-DAgger}}} & \multicolumn{4}{c}{\textbf{\textsc{DAgger}}} & \\\cmidrule(lr){3-8}\cmidrule(lr){9-12}
  
    \phantom{a} & \phantom{a} & \multicolumn{2}{c}{\textbf{PI-Net}} & \multicolumn{2}{c}{\textbf{RNN}} & \multicolumn{2}{c}{\textbf{FNN}} & \multicolumn{2}{c}{\textbf{RNN}} & \multicolumn{2}{c}{\textbf{FNN}} & \multicolumn{2}{c}{\textbf{Expert}} \\
    
    \textbf{Task} & \textbf{Params} & \% & Cost & \% & Cost & \% & Cost &\% & Cost &\% & Cost &\% & Cost\\\midrule
    
    \multirow{4}{*}{\shortstack[c]{Quad-\\copter\\Circle}} 
    & baseline 	   &$100$&$55\pm 16$  	&$100$&$9\pm 1$ &$100$&$9\pm 1$    &$100$&$8\pm 1$ 	&$\mb{100}$&$8\pm  0$  &$100$&$8\pm 0$\\
    & $\sigma+0.9$ &$100$&$47\pm 13$ &$100$&$13\pm 3$    &$\mb{100}$&$12\pm 2$   &$96.9$&$12\pm 2$ &$79.7$&$10\pm 2$ &$100$&$9\pm 0$\\
    & $x_0+0.5$    &$60.9$&$59\pm 15$ &$96.0$&$23\pm 11$   &$\mb{100}$&$21\pm 12$  &$89.8$&$20\pm 9$ &$50.0$&$10\pm 2$   &$100$&$9\pm 0$\\
    & $l\div 2$    &$99.2$&$46\pm 12$ &$\mb{100}$&$9\pm 1$    &$\mb{100}$&$9\pm 1$	   &$89.8$&$8\pm 1$  &$93.0$&$89\pm 3$ 	&$100$&$6\pm 0$\\
    & $m+0.3$      &$\mb{99.2}$&$41\pm 5$ &$74.2$&$38\pm 5$    &$40$&$72\pm 14$   &$0$&- 			&$0$&- 		 		&$100$&$22\pm 0$\\
    \\
    \multirow{4}{*}{\shortstack[c]{Quad-\\copter\\Fig-8}}
    & baseline 	   &$100$&$10\pm 1$   	&$\mb{100}$&$8\pm 1$    &$100$&$9\pm 5$    &$100$&$9\pm 1$     &$100$&$9\pm 1$   &$100$&$8\pm 0$\\
    & $\sigma+0.9$ &$\mb{96.7}$&$13\pm 7$   &$92.2$&$13\pm 5$  &$93.8$&$23\pm 20$ &$100$&$9\pm 1$     &$7.8$&$38\pm 25$  &$100$&$8\pm 0$\\
    & $x_0+0.5$    &$66.0$&$12\pm 2$   &$\mb{100}$&$13\pm 6$   &$93.0$&$16\pm 10$ &$93.8$&$15\pm 11$  &$14.0$&$10\pm 4$  &$100$&$9\pm 0$\\
    & $l\div 2$    &$\mb{100}$&$9\pm 1$ 		&$100$&$9\pm 2$    &$98$&$9\pm 1$     &$100$&$10\pm 1$    &$86.7$&$9\pm 0$  &$100$&$6\pm 0$\\
    & $m+0.3$      &$\mb{97.5}$&$39\pm 3$   &$14.8$&$37\pm 21$ &$68.8$&$83\pm 8$  &$0$&- 				&$0$&-	 		   &$100$&$22\pm 0$\\
    \\
    \multirow{3}{*}{\shortstack[c]{Cart-\\Pole}}
    & baseline 	   &$92.1$&$13\pm 18$ &$81.8$&$16\pm 19$   &$\mb{99.0}$&$\mb{1\pm 2}$   &$99.0$&$6\pm 4$ &$92.8$&$4\pm 7$   &$100$&$1\pm 1$\\
    & $\sigma+0.9$ &$78.1$&$38\pm 23$ &$65.8$&$48\pm 24$   &$\mb{88.5}$&$\mb{41\pm 24}$ &$77.0$&$45\pm 24$ &$76.8$&$42\pm 25$ &$99.2$&$4\pm 3$\\
    & $M+.2$      &$\mb{92.2}$&$\mb{16\pm 20}$ &$26.8$&$39\pm 29$   &$71.5$&$10\pm 15$ &$51.8$&$32\pm 26$ &$35.4$&$12\pm 17$ &$99.8$&$6\pm 6$\\
    & $l+0.2$      &$\mb{90.6}$&$\mb{15\pm 18}$ &$12.7$&$41\pm 28$   &$47.5$&$7\pm 10$  &$33.0$&$13\pm 20$ &$11.7$&$9\pm 15$ &$100$&$3\pm 4$\\
    \bottomrule
  \end{tabular}
  \end{center}
 \label{tab:1}
\end{table*}

\subsection{Results and Discussion}
The main results of the various tasks and policies are reported in Table \ref{tab:1}.  In each task we report the percentage of successful task completions and the average loss accumulated by the successful trajectories only.  We highlight policies with the highest success rates, with ties broken by the lowest average loss.  We also show the performance of the MPC expert on each of the tasks.  The expert with a fixed model is able to fully generalize when applied to models with different parameters, which demonstrates the generalization capacity of MPC methods.  Not shown in our results are those of policies trained with simple supervised learning on expert data.  These policies completely failed to generalize on any of the tasks.

\textbf{MPC-type policies show improved generalization over non-MPC policies.}  When evaluated for varying noise, initial conditions, and system dynamics, MPC-type policies trained with \textsc{MPC-DAgger} consistently perform better.  Figure \ref{fig:fnn_quad} shows an example comparing a reactive policy and an RNN policy trained with \textsc{MPC-DAgger}, where spread of the initial state distribution was increased.  The RNN policy is able to generalize to these new initial states and the trajectories quickly converge correctly, whereas the reactive policy is brittle to these perturbations and many trajectories fail catastrophically.  

\textbf{PI-Net sucessfully learns a policy that improves generalization.}
Figures \ref{fig:pinets_figof8} and \ref{fig:cartpole_pinets} shows a comparison of the performance of PI-Net with the other policies under varying parameters.  Other policies are less able to generalize to the changing task; this is especially true for the reactive FNN policy which is particularly brittle.  Although in these figures we showcase the best performance of PI-Net, Table \ref{tab:1} shows overall good performance of PI-Net relative to the other policies in general.  We believe that further improvements can be made with a deeper exploration of the design space.

\textbf{RNN policies trained with Vanilla \textsc{DAgger} improve performance but remain inadequate.}  
Although training an RNN policy to produce sequences of controls given sequences of states improves generalization performance when compared to a simple FNN reactive policy, it generally does not perform as well as an MPC-type policy.  This supports our hypothesis that MPC-type policies, which iteratively optimize sequences of controls, are better able to withstand modeling error and disturbances.

\section{Conclusion} 
\label{sec:conclusion}
MPC-inspired policies are recurrent neural networks which iteratively reoptimize sequences of controls.  We have shown that MPC-type policies are more robust to disturbances and generalize better than their reactive counterparts.  We propose a simple extension to the \textsc{DAgger} algorithm and show how it can be used to efficiently train such policies.
PI-Net is a planning policy which acts in an MPC fashion.  It is a sampling based optimal controller which consists of dynamics rollouts, a cost function, and an update rule to reoptimize a given control sequence.  We have shown that our extension to \textsc{DAgger} allows efficient and scalable training of PI-Nets.  PI-Net generalizes well in comparison with other policies.
There are many directions for future work.  We believe that there is a potentially very large design space for PI-Net that can be explored, e.g. methods to reduce the number of rollouts, etc.  Training and testing PI-Net is still computationally difficult and improvements to efficiency will make design faster.  Other future directions include training MPC-type policies with reinforcement learning, and investigating better training methodologies, drawing from the sequence prediction community.  We believe that this work is an important step forward towards learning efficient and scalable policies with planning architectures for continuous state and action spaces.

\bibliography{paper_template}
\bibliographystyle{icml2018}

\end{document}